\documentclass[12pt]{article}
\usepackage{url}
\usepackage[centertags]{amsmath}
\usepackage{algorithm}
\usepackage{algpseudocode}
\usepackage{amsfonts}
\usepackage{graphicx}
\usepackage[export]{adjustbox}

\title{Quasi Manhattan Wasserstein Distance}
\author{Evan Unit Lim\thanks{Email: elim@ntnu.edu.tw}} 
\date{\today}

\begin{document}
\maketitle

\begin{abstract}
The Quasi Manhattan Wasserstein Distance (QMWD) is a metric designed to quantify the dissimilarity between two matrices by combining elements of the Wasserstein Distance with specific transformations. It offers improved time and space complexity compared to the Manhattan Wasserstein Distance (MWD) while maintaining accuracy. QMWD is particularly advantageous for large datasets or situations with limited computational resources. This article provides a detailed explanation of QMWD, its computation, complexity analysis, and comparisons with WD and MWD.
\end{abstract}

\section{Introduction} 
The Wasserstein distance, also known as the earth mover's distance, is a measure of similarity between probability distributions on a metric space. It can be used to compare images by computing the minimum cost of transferring one image into another. One compelling application of Wasserstein distance is the Wasserstein Generative Adversarial Network (WGAN), which is a variant of GAN that aims to improve the stability of learning and provide meaningful learning curves~\cite{arjovsky2017wasserstein}. However, solving the optimal transport problem in 2D is computationally inefficient. Hence, data are often vectorized (flattened) into 1D arrays prior to the computation of Wasserstein distances~\cite{arjovsky2017wasserstein}. As a trade off, this overpenalizes vertical divergences. We propose a novel metric that has the same overall time complexity as computing Wasserstein distances on vectorized arrays but with elevated accuracy.\footnote{Example code available at \url{https://github.com/evlim/qmwd}}

\section{Manhattan Distance}
The Manhattan distance, also known as the L1 distance or taxicab distance, between two points \(A(x_1, y_1)\) and \(B(x_2, y_2)\) in a two-dimensional space is defined as:

\[
\text{MD}(A, B) = |x_2 - x_1| + |y_2 - y_1|
\]
where \(|x_2 - x_1|\) represents the absolute difference in the x-coordinates, and \(|y_2 - y_1|\) represents the absolute difference in the y-coordinates. It measures the sum of the horizontal and vertical distances required to move from point \(A\) to point \(B\) along grid lines, resembling the path a taxi might take in a city~\cite{black2006manhattan}.

\section{Wasserstein Distance}
The Wasserstein distance, also known as the Earth Mover's distance or Kantorovich-Rubinstein metric, is a metric that quantifies the minimum "work" or cost needed to transform one probability distribution \(P(x, y)\) into another distribution \(Q(x', y')\). It is defined as
\[
\text{WD}(P, Q) = \min_{\gamma} \sum_{x'} \sum_{y'} c(x, y, x', y') \gamma(x, y)
\]
where \(\gamma(x, y)\) represents the transportation plan that specifies how much mass is to be moved from point \(P(x, y)\) to point \(Q(x', y')\). \(c(x, y, x', y')\) is the cost or distance between source point \((x, y)\) and destination point \((x', y')\)~\cite{vaserstein1969markov}\cite{kantorovich1960mathematical}\cite{rubner2000earth}. The choice of this cost function determines the specific Wasserstein distance variant (e.g., Euclidean, Manhattan, etc.).

\section{Manhattan Wasserstein Distance}
\subsection{Definition}
The Manhattan Wasserstein distance is defined as
$$\text{MWD}(P, Q) = \min_{\gamma} \sum_{x'} \sum_{y'} \left( |x' - x| + |y' - y| \right) \gamma(x, y)$$
where \(x'\) and \(y'\) represent the coordinates in the matrix grid, \(\gamma(x, y)\) represents the transportation plan that determines the amount of "mass" to be moved from \(P(x, y)\) to \(Q(x', y')\). \(|x' - x| + |y' - y|\) is the Manhattan distance between the source coordinates \((x, y)\) and the destination coordinates \((x', y')\).

In short, the Manhattan distance is used as the cost function to calculate the Wasserstein distance between the two matrices \(P\) and \(Q\), which have the same sum of their matrix elements. A pseudocode of MWD is shown in Algorithm~\ref{alg:manhattan_wasserstein}.

\begin{algorithm}[H]
\caption{Manhattan Wasserstein Distance}
\label{alg:manhattan_wasserstein}
\begin{algorithmic}[1]
\Require Source matrix $P$ of size $m \times n$ and destination matrix $Q$ of the same size.
\Ensure Manhattan Wasserstein distance between $P$ and $Q$.

\State Compute the cost matrix $cost\_matrix$ of size $m \times n \times m \times n$.
\For{$i = 0$ to $m-1$}
    \For{$j = 0$ to $n-1$}
        \For{$k = 0$ to $m-1$}
            \For{$l = 0$ to $n-1$}
                \State $cost\_matrix[i][j][k][l] \gets |i - k| + |j - l|$
            \EndFor
        \EndFor
    \EndFor
\EndFor

\State Initialize transportation plan matrix $\gamma$ of size $m \times n \times m \times n$.
\State Define the linear programming problem:
\Statex $\text{objective} = \text{Minimize} \left( \sum_{i=0}^{m-1} \sum_{j=0}^{n-1} \sum_{k=0}^{m-1} \sum_{l=0}^{n-1} cost\_matrix[i][j][k][l] \cdot \gamma[i][j][k][l] \right)$

\State Add constraints:
\Statex $\gamma \geq 0$
\Statex $\sum_{j=0}^{n-1} \gamma[i][j][k][l] = P[i][j]$ for all $i, k, l$
\Statex $\sum_{k=0}^{m-1} \gamma[i][j][k][l] = Q[i][j]$ for all $i, j, l$

\State Solve the linear programming problem to find optimal $\gamma$.

\State \Return MWD
\end{algorithmic}
\end{algorithm}

\section{Quasi Manhattan Wasserstein Distance}
\subsection{Definition}
In this section, we propose the Quasi Manhattan Wasserstein Distance (QMWD). It is designed to quantify the dissimilarity between two matrices with identical sum of their matrix elements, $P$ and $Q$, by combining elements of the Wasserstein Distance (WD) with specific transformations. QMWD is defined as follows:

\begin{align*}
\text{QMWD}(P, Q) = \max\Bigg(&\frac{\text{WD}(\text{vec}(P), \text{vec}(Q))}{n} \\
&+ \text{WD}(\text{vec}(P), \text{vec}(Q)) \mod n, \\
&\frac{\text{WD}(\text{vec}(R(P)), \text{vec}(R(Q)))}{m} \\
&+ \text{WD}(\text{vec}(R(P)), \text{vec}(R(Q))) \mod m, \\
&\frac{\text{WD}(\text{vec}(P^T), \text{vec}(Q^T))}{m} \\
&+ \text{WD}(\text{vec}(P^T), \text{vec}(Q^T)) \mod m\Bigg)
\end{align*}
where $P$ and $Q$ are matrices of size $m \times n$. vec(P) and vec(Q) represent vectorized versions of the input arrays \(P\) and \(Q\), respectively, in row-major order. $R(P)$ and $R(Q)$ denote the 90 degree rotation of matrices $P$ and $Q$. $P^T$ and $Q^T$ are the transposes of matrices $P$ and $Q$.

QMWD combines the WD between vectorized matrices, the WD between rotated matrices, and the WD between transposed matrices, each scaled by the dimensions $n$ and $m$, and modified by the modulo operation. A pseudocode of QMWD is shown in Algorithm~\ref{alg:quasi_manhattan_wasserstein}. It's important to note that this method assumes positive integer elements. Additional scaling is required for cases where matrices P and Q have different sums of matrix elements or when addressing floating-point numbers. For those cases, please refer to the code available at \url{https://github.com/evlim/qmwd}.

\begin{algorithm}[H]
\caption{Quasi Manhattan Wasserstein Distance}
\label{alg:quasi_manhattan_wasserstein}
\begin{algorithmic}[1]
\Require Source matrix $P$ of size $m \times n$ and destination matrix $Q$ of the same size.
\Ensure Quasi Manhattan Wasserstein Distance between $P$ and $Q$.

\Function{QMWD}{$P, Q$}
    \State $WD1 \gets \text{WD}(\text{vec}(P), \text{vec}(Q))$
    \State $WD2 \gets \text{WD}(\text{vec}(R(P)), \text{vec}(R(Q)))$
    \State $WD3 \gets \text{WD}(\text{vec}(P^T), \text{vec}(Q^T))$
    \State $QMWD \gets \max\left(\frac{WD1}{n} + WD1 \mod n, \frac{WD2}{m} + WD2 \mod m, \frac{WD3}{m} + WD3 \mod m\right)$
    \State \Return $QMWD$
\EndFunction
\end{algorithmic}
\end{algorithm}

\section{Complexity Analysis}
\subsection{Manhattan Wasserstein Distance (MWD)}
The computational complexity of the Manhattan Wasserstein Distance (MWD) can be analyzed as follows:

\subsubsection{Cost Matrix Computation}
The first step in calculating MWD is to compute the cost matrix of size $m \times n \times m \times n$. This involves four nested loops, each ranging from $0$ to $m-1$ or $n-1$, resulting in a time complexity of $O(m^2n^2)$.

\subsubsection{Linear Programming}
The linear programming problem to find the optimal transportation plan $\gamma$ involves solving a system of linear equations. In practice, various optimization algorithms can be employed, such as the simplex method or interior-point methods. The worst-case time complexity of solving linear programming problems can be exponential, but in many practical scenarios, it converges efficiently. Let $T$ be the time complexity for solving the linear programming problem, which depends on the specific solver used. The overall time complexity for this step can be expressed as $O(T)$.

\subsubsection{Overall Complexity}
The dominant factor in the overall complexity of MWD is the computation of the cost matrix, resulting in a final time complexity of $O(m^2n^2) + O(T)$.

\subsection{Quasi Manhattan Wasserstein Distance (QMWD)}
The computational complexity of the Quasi Manhattan Wasserstein Distance (QMWD) is analyzed as follows:

\subsubsection{Vectorization}
The vectorization of matrices $P$ and $Q$ involves flattening them, which can be done in $O(mn)$ time.

\subsubsection{Wasserstein Distance (WD) Calculation}
QMWD computes three WD values between 1D arrays representing vectorized matrices, rotated matrices, and transposed matrices. Each of these WD computations has a time complexity of $O(mn) + O(T)$, as discussed in the WD section.

\subsubsection{Overall Complexity}
The overall time complexity of QMWD is dominated by the three WD calculations. Therefore, the final time complexity of QMWD is $3 \times \left(O(mn) + O(T)\right) = O(3mn) + O(3T) = O(mn) + O(T)$.

\subsection{Simulation}
The simulation involved the computation of three distance metrics: WD(vec($P$), vec($Q$)), MWD($P$, $Q$), and QMWD($P$, $Q$). In this study, matrices $P$ and $Q$ were randomly generated, with dimensions $m \times n$, where $n$ was fixed at 30, and $m$ varied from 2 to 30. For each value of $m$, we conducted 20 simulations using randomly generated $P$ and $Q$. 

To assess the quality of the calculated metrics, we computed the relative errors for both WD and QMWD compared to MWD. MWD was considered as the "ground truth" metric for comparison. The error for WD was determined as $|\text{MWD}(P, Q) - \text{WD}(\text{vec}(P), \text{vec}(Q))| / \text{MWD}(P, Q)$, while the error for QMWD was calculated as $|\text{MWD}(P, Q) - \text{QMWD}(P, Q)| / \text{MWD}(P, Q)$. Additionally, we measured the execution time to assess computational efficiency.

\begin{figure}[H]
    \centerline{\includegraphics[width=15cm]{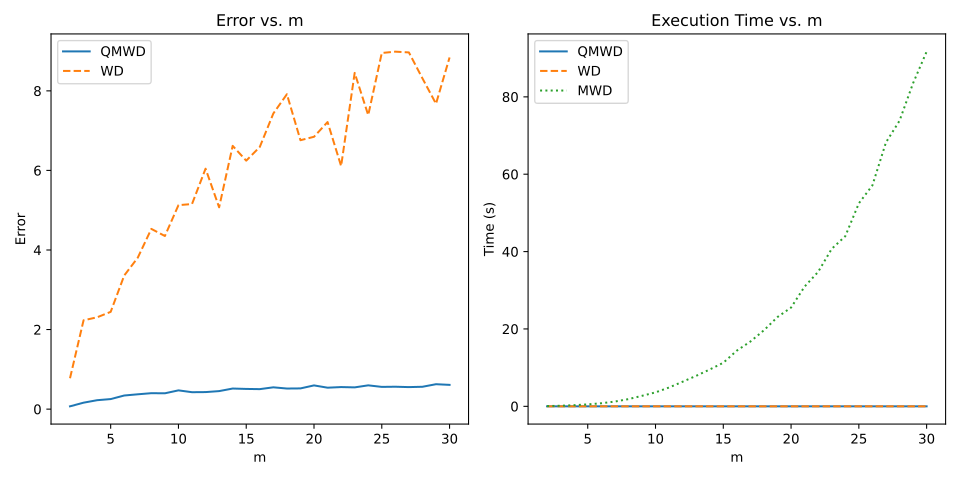}}
    \caption{Plots of Accuracy and Execution Time (n=30)}\label{fig:average_accuracy_and_execution_time_plots}
\end{figure}

The results of these simulations are presented in Figure \ref{fig:average_accuracy_and_execution_time_plots}.

\subsection{Remarks}
The time complexity of both MWD and QMWD depends on the size of the input matrices ($m$ and $n$) and the efficiency of the linear programming solver used ($T$). In practice, the choice of solver and the sparsity of the cost matrix can significantly affect the actual runtime.

It's important to note that the complexity analysis provided here assumes dense matrices and a general linear programming solver. Specialized solvers or matrix structures can lead to optimizations that improve practical performance.

\section{Advantages and Disadvantages}
\subsection{Improved Time and Space Complexity}
One of the key advantages of Quasi Manhattan Wasserstein Distance (QMWD) over Manhattan Wasserstein Distance (MWD) is its significantly improved time and space complexity. QMWD's time complexity is typically linear or near-linear with respect to the dimensions of the input matrices. In contrast, MWD's time complexity is quadratic, as it needs to compute the cost matrix of size $m \times n \times m \times n$, which also results in higher space complexity. This makes QMWD computationally more efficient, especially for large datasets or situations where computational resources are limited.

\subsection{Absence of 2D Transportation Plan}
Unlike MWD, QMWD does not compute the 2D transportation plan that specifies how mass is moved from one point to another in the input matrices. This characteristic may limit the applicability of QMWD in cases where the transportation plan itself is of interest or significance.

\section*{Acknowledgement}
This project was inspired by a class project from 'CS 7637: Knowledge-Based Artificial Intelligence - Cognitive Systems' at the Georgia Institute of Technology in Summer 2023. The class project aimed to create an AI agent that can pass a human intelligence test. This project is supported by NSTC 112-2314-B-001-010 grant titled 'Development of Deep Generative Models for Synthesizing Genomic Data at Scale to Improve Disease Risk Prediction' from the National Science and Technology Council. The grant was acquired by Evan Unit Lim, Aylwin Ming Wee Lim, and Cathy SJ Fann. Evan Unit Lim is also supported by the Institute of Biomedical Sciences at Academia Sinica, and the Computational Thinking and Programming Education Division at National Taiwan Normal University.

\bibliography{bibliography.bib}
\end{document}